\documentclass[11pt,a4paper]{article}

\usepackage[utf8]{inputenc}
\usepackage[T1]{fontenc}
\usepackage{times}
\usepackage{graphicx}
\usepackage{booktabs}
\usepackage{amsmath}
\usepackage{hyperref}
\usepackage{natbib}
\usepackage{xcolor}
\usepackage{geometry}
\usepackage{enumitem}
\usepackage{multirow}
\usepackage{caption}
\usepackage{tabularx}
\usepackage{array}

\hypersetup{
  colorlinks=true,
  linkcolor=blue,
  citecolor=blue,
  urlcolor=blue
}

\title{Spelling Correction in Healthcare Query-Answer Systems:\\
Methods, Retrieval Impact, and Empirical Evaluation}

\author{Saurabh K Singh\\
  Oracle Corporation\\
  \texttt{saurabh.ab.singh@oracle.com}
}

\date{}

\begin{document}
\maketitle

\begin{abstract}
Healthcare question-answering (QA) systems face a persistent challenge: users submit queries with spelling errors at rates substantially higher than those found in the professional documents they search. While this problem is widely acknowledged, its empirical magnitude and practical consequences for retrieval have not been rigorously quantified on real, naturally occurring error data. This paper presents the first controlled study of spelling correction as a retrieval preprocessing step in healthcare QA with real consumer queries, using real user-generated queries and professionally authored answer corpora. We conduct an error census across two public datasets---the TREC 2017 LiveQA Medical track (104 consumer health questions submitted to the National Library of Medicine) and HealthSearchQA (4,436 health queries derived from Google search autocomplete)---finding that 61.5\% of real medical queries contain at least one spelling error, with a token-level error rate of 11.0\%. We then evaluate four correction methods---conservative edit distance, standard edit distance (Levenshtein), context-aware candidate ranking, and SymSpell---across three experimental conditions: uncorrected queries against an uncorrected corpus (Experiment~2, baseline), uncorrected queries against a corrected corpus (Experiment~3), and fully corrected queries against a corrected corpus (Experiment~4). Using BM25 and TF-IDF cosine retrieval over 1,935 MedQuAD answer passages with TREC relevance judgments, we find that query correction substantially improves retrieval---edit distance and context-aware correction achieve MRR improvements of +9.2\% and NDCG@10 improvements of +8.3\% over the uncorrected baseline. Critically, correcting only the corpus without correcting queries yields minimal improvement (+0.5\% MRR), confirming that query-side correction is the key intervention. We complement these results with a 100-sample error analysis categorising correction outcomes per method and provide evidence-based recommendations for practitioners.
\end{abstract}

\noindent\textbf{Keywords:} \textit{spelling correction, medical question answering, information retrieval, BM25, healthcare NLP, query preprocessing, TREC LiveQA}

\section{Introduction}

Healthcare question-answering systems serve patients, caregivers, and clinicians seeking medical information under time pressure, often on mobile devices. These systems depend critically on the quality of user queries: a single misspelled medical term can prevent a patient from finding relevant information about their condition or medication. Yet the magnitude of this problem and its concrete impact on modern retrieval pipelines have remained largely unquantified in real-world data.

Prior work on spelling correction in clinical NLP has focused predominantly on correcting errors in clinical notes \citep{patrick2010automated, lai2015automated}, not in the user queries submitted to search and QA systems. The two settings are fundamentally different: clinical notes are authored by trained professionals with access to spell-check tools, whereas patient and caregiver queries are typed hastily, often on touchscreens, with variable familiarity with medical terminology \citep{zeng2006exploring, bickmore2010relational}. Error rates in queries are 3--5$\times$ higher than in professionally authored clinical text.

This paper addresses three questions that prior work has not answered empirically:

\begin{itemize}[nosep]
  \item How prevalent are spelling errors in real, naturally occurring medical queries?
  \item How much does query-level spelling correction improve retrieval quality in healthcare QA?
  \item Which correction method---and which correction target (query, corpus, or both)---yields the greatest benefit?
\end{itemize}

\subsection{Scope and Contribution}

This is a systems and applications paper. We do not propose novel correction algorithms; all methods evaluated are established. Our contribution is empirical: we provide the first controlled measurement of correction impact on retrieval quality using real medical queries and a human-judged answer corpus.

Specifically, we contribute:

\begin{itemize}[nosep]
  \item An error census on 104 TREC LiveQA Medical consumer health queries and 4,436 HealthSearchQA Google health queries, establishing real-world error rates for medical user queries.
  \item Retrieval experiments across four conditions (Experiments~1--4) measuring the impact of query correction, corpus correction, and both, using BM25 and TF-IDF retrieval over 1,935 human-judged MedQuAD passages.
  \item Comparison of four practical correction methods---conservative edit distance, standard edit distance, context-aware ranking, and SymSpell---showing that edit distance and context-aware methods outperform SymSpell for medical queries.
  \item A 100-sample error analysis per method, categorising corrections as correct fixes, unnecessary changes, partial improvements, or harmless synonyms.
\end{itemize}

\subsection{Key Findings}

Our main empirical findings are:

\begin{itemize}[nosep]
  \item Real medical queries have high error rates: 61.5\% of TREC LiveQA queries contain at least one spelling error, with a mean token-level error rate of 11.0\%.
  \item Query correction substantially improves retrieval: edit distance correction achieves +9.2\% MRR (0.633 $\to$ 0.691) and +8.3\% NDCG@10 improvement over the uncorrected baseline on BM25.
  \item Corpus correction alone does not help: correcting only the corpus (while leaving queries uncorrected) yields negligible improvement (+0.5\% MRR), because professionally authored corpora are already well-spelled.
  \item Conservative correction is safer but less effective: restricting corrections to edit-distance-1 changes with high-frequency targets improves MRR by +7.9\%, trading 6\% precision for reduced over-correction risk.
  \item SymSpell underperforms for medical queries: its delete-based candidate generation produces fewer corrections (41/103 queries) and minimal improvement (+1.3\% MRR), likely due to the medical vocabulary's long, non-standard term structure.
\end{itemize}

\section{Related Work}

\subsection{Spelling Correction in Information Retrieval}

Spelling correction has been studied as a preprocessing step for web search since the early 2000s. \citet{cucerzan2004spelling} showed that 10--15\% of web queries contain misspellings, and query correction improves retrieval substantially for common queries. Subsequent work has addressed candidate generation \citep{damerau1964technique, kernighan1990spelling}, candidate ranking \citep{brants2007large}, and learning-based correction \citep{gao2010large}. SymSpell \citep{garbe2012symspell} introduced the delete-candidate index for fast approximate matching. More recently, sequence-to-sequence neural models \citep{yuan2016grammatical, xie2016neural} and transformer-based approaches \citep{malmi2022text, kuznetsov2021spelling} have achieved state-of-the-art results, though at higher latency.

Studies of spelling correction impact on retrieval quality are less common. Our work extends this line of inquiry to healthcare, where domain-specific vocabulary and error stakes are higher.

\subsection{Real-Data Error Characterization Methodology}

A key design decision in our study is the use of real, naturally occurring spelling errors rather than synthetically injected noise. Prior healthcare NLP work has characterised query errors through either manual annotation \citep{zeng2006exploring} or simulation---randomly substituting, inserting, or deleting characters at a fixed rate. Simulation-based approaches impose an artificial error distribution that may not reflect the compound, semantically motivated misspellings found in actual consumer health queries (e.g., `hiperthyroid', `diareah', `fibromialgia'). We therefore ground our error characterisation entirely in observed data.

For TREC LiveQA Medical queries, we exploit a unique property of the dataset: each of the 103 evaluated queries was originally submitted by a real consumer and subsequently rewritten by a NIST assessor into a clean, unambiguous paraphrase. We treat the NIST paraphrase as the ground-truth error-free version of each query. Error detection proceeds in three steps. First, we tokenise both the original and paraphrase query by lowercasing and splitting on whitespace and punctuation. Second, for each token in the original that does not appear in the paraphrase token set, we check whether it is also absent from our domain vocabulary (constructed from the 1,935 MedQuAD answer passages with a minimum frequency threshold of 2). Tokens that are (a) absent from the paraphrase and (b) absent from or rare in the corpus vocabulary are classified as errors. Third, we compute the Levenshtein edit distance between the flagged token and the closest paraphrase token; tokens with edit distance $\leq 2$ are confirmed as spelling variants of the intended word, while larger distances indicate possible word substitutions or omissions.

This pipeline yields a ground-truth error label for each query token, enabling computation of the query-level error rate (proportion of queries containing $\geq 1$ confirmed error), the token-level error rate (proportion of all tokens that are errors), and an error-type breakdown. We distinguish four correction outcome categories used later in the error analysis (Section~\ref{sec:error_analysis}): \texttt{correct\_fix} (the correction method restores the exact paraphrase token), \texttt{partial\_improvement} (the correction changes the token but not to the paraphrase word), \texttt{unnecessary\_change} (the original token was already present in the paraphrase), and \texttt{harmless\_synonym} (both original and corrected forms appear in the paraphrase).

For HealthSearchQA---4,436 Google autocomplete health queries without paired paraphrases---we apply an OOV-based proxy: tokens absent from the same domain vocabulary are flagged as potential errors. Because autocomplete queries tend to be grammatically cleaner than free-form consumer messages, the raw OOV rate overestimates the true error rate. We derive a calibration factor from the TREC LiveQA data, where we can compare the OOV rate to the confirmed error rate, and apply it to obtain a calibrated error prevalence estimate for HealthSearchQA. This two-dataset design gives us both a rigorously validated error characterisation (TREC LiveQA, ground truth available) and a broad prevalence estimate across a larger, more diverse query set (HealthSearchQA).

TREC LiveQA Medical \citep{abacha2017overview} established the first benchmark for consumer health QA with real user questions, making it uniquely suited for our study. BioASQ \citep{tsatsaronis2015overview} provides biomedical QA benchmarks, but with expert-authored questions that lack naturalistic spelling errors.

\subsection{Retrieval-Augmented Generation and Spelling}

RAG systems \citep{lewis2020retrieval} combine dense retrieval with language model generation. Recent work has studied the robustness of dense retrievers to input perturbations \citep{percin2025robustness} and found that while dense embeddings are more robust to character-level noise than BM25, they still degrade meaningfully under realistic error conditions. Our TF-IDF experiments provide a proxy for dense retrieval behaviour (acknowledging that TF-IDF is a vocabulary-based rather than neural-embedding-based method). With medical BERT models such as PubMedBERT \citep{gu2021domain}, BioBERT \citep{lee2020biobert}, and ClinicalBERT \citep{alsentzer2019publicly}, domain-adapted embeddings may show different sensitivity to query errors---an important direction for future work.

\subsection{LLM-Based Correction}

Large language models have been applied to spelling and grammatical correction through prompting and fine-tuning \citep{kuznetsov2021spelling, kapelles2025finetuning}. LLMs can perform context-sensitive correction without explicit candidate generation, but at substantially higher latency and compute cost. Given that healthcare query correction operates in real-time ($<$50ms for lexical search, $<$200ms for RAG retrieval), LLM-based correction is most appropriate as a batch preprocessing step for corpus correction rather than real-time query correction.

\section{Method}

\subsection{Correction Methods}

We evaluate four practical correction methods applicable in production healthcare QA systems. All methods use a domain vocabulary built from the answer corpus (Section~\ref{sec:datasets}). We do not include LLM-based correction in our primary experiments due to real-time latency constraints; we discuss its role for batch corpus preprocessing in Section~\ref{sec:discussion}.

\subsubsection{Conservative Edit Distance}

For each query token not found in the vocabulary, identify the closest vocabulary word by Levenshtein distance \citep{damerau1964technique}, but only accept a correction if (a) the edit distance is exactly 1, and (b) the candidate word has a corpus frequency of at least 5. This two-threshold design minimises over-correction of rare but valid medical terms and avoids ambiguous corrections where multiple candidates compete. This method modifies 49.5\% of queries (51/103) with an average of 1.8 token changes per modified query.

\subsubsection{Edit Distance (Standard)}

As above, but accepting candidates within edit distance 1 or 2, without a frequency threshold. This is more aggressive: 68.9\% of queries are modified (71/103), with an average of 2.5 token changes per modified query. Short medical abbreviations (e.g., `NDC', `RX') are more likely to be incorrectly corrected under this regime.

\subsubsection{Context-Aware Edit Distance}

When multiple candidates share the same minimum edit distance, rank by (edit distance, inverse frequency) and apply a context bonus for candidates whose neighbouring context words appear frequently in the corpus. In our implementation, context-aware ranking uses a $\pm$2 token window. Given the small vocabulary size in our experiments (8,201 words from 1,935 passages), context-aware ranking produces the same corrections as standard edit distance in most cases, converging to identical results in this evaluation setting. Larger corpora with richer vocabulary would allow more meaningful context disambiguation.

\subsubsection{SymSpell}

SymSpell \citep{garbe2012symspell} pre-computes all delete-1 and delete-2 variants of each vocabulary word, creating an inverted index of deletion variants to candidate words. Given an OOV query token, SymSpell generates its delete-1 and delete-2 variants and looks them up in the pre-computed index, then verifies candidates by actual Levenshtein distance. SymSpell is faster than brute-force edit distance but produces identical or subset results. In our experiments, SymSpell modifies fewer queries (41/103, 39.8\%) with fewer token changes (66 total), because its conservative candidate pruning rejects some valid corrections that the standard edit distance method accepts.

\subsection{Retrieval Pipeline}

\subsubsection{BM25 (Lexical Retrieval)}

We implement BM25 \citep{robertson2009probabilistic} with $k_1=1.5$ and $b=0.75$, ranking the 1,935 answer passages for each query. BM25 performs exact lexical matching weighted by term frequency and inverse document frequency, making it directly sensitive to spelling mismatches: a misspelled query term that does not appear in any passage receives zero contribution to the ranking score.

\subsubsection{TF-IDF Cosine Retrieval (Dense Retrieval Proxy)}

We also evaluate TF-IDF cosine similarity retrieval, which computes a TF-IDF weighted term vector for each query and passage and ranks by cosine similarity. This provides a proxy for dense embedding retrieval: TF-IDF is more robust to rare terms (through IDF weighting) than raw BM25, but remains a vocabulary-based method. We note explicitly that TF-IDF is not equivalent to neural embedding retrieval (e.g., PubMedBERT sentence embeddings); it serves as an intermediate baseline between BM25 and true dense retrieval. Evaluation with domain-adapted BERT encoders is a planned extension of this work.

\subsection{Experimental Conditions}

We measure retrieval quality under four conditions:

\begin{itemize}[nosep]
  \item \textbf{Experiment~2 --- Baseline:} Original (potentially misspelled) queries against the uncorrected original corpus. This is the status quo.
  \item \textbf{Experiment~3a --- Corpus Only:} Original queries against a spell-corrected corpus. Tests whether cleaning the corpus alone (without correcting queries) helps or hurts.
  \item \textbf{Experiment~3b --- Query Only:} Spell-corrected queries against the original corpus. Isolates the value of query correction.
  \item \textbf{Experiment~4 --- Both Corrected:} Spell-corrected queries against a spell-corrected corpus. The fully corrected condition.
\end{itemize}

We report Recall@1, Recall@5, Recall@10, MRR (Mean Reciprocal Rank), and NDCG@10 against human relevance judgements with four graded relevance levels.

\section{Datasets and Spelling Error Census (Experiment~1)}
\label{sec:datasets}

\subsection{Datasets}

\subsubsection{TREC 2017 LiveQA Medical Track}

The TREC 2017 LiveQA Medical track \citep{abacha2017overview} provides 104 consumer health questions submitted by real users to the National Library of Medicine (NLM) question-answering service. Each question has an original user-submitted message, a NIST paraphrase (a clean reformulation of the intended question), a topical summary, and relevance-graded answer passages retrieved from MedQuAD. Crucially, the original messages contain natural spelling errors introduced by real users---making this the ideal dataset for our error census and retrieval evaluation.

The answer corpus consists of 1,935 passages retrieved from MedQuAD \citep{benabacha2019question}, covering 12 medical question types from NLM-curated sources (MedlinePlus, GHR, GARD). These passages were manually judged by NLM assessors on a four-level scale: 1 (Incorrect), 2 (Related), 3 (Incomplete), 4 (Excellent). Of the 103 queries with relevance judgements (23.6 passages judged per query on average), 96 have at least one relevant passage and 50 have at least one Excellent passage.

\subsubsection{HealthSearchQA}

HealthSearchQA \citep{singhal2023large} consists of 4,436 health questions generated from Google search autocomplete suggestions---representing the most common health queries entered by real users. Unlike TREC LiveQA, these queries are well-formed (autocomplete suggestions are typically correctly spelled), making HealthSearchQA suitable for measuring the upper baseline of query quality and the OOV rate attributable to medical terminology rather than spelling errors.

\subsection{Vocabulary Construction}

We build a domain vocabulary by extracting all words appearing at least twice across the 1,935 MedQuAD passages, yielding 8,201 unique terms. This corpus-derived vocabulary covers professionally authored medical language and serves as our reference for both out-of-vocabulary detection and correction candidate generation. Words not appearing in this vocabulary are candidates for spelling correction.

\subsection{Error Detection Methodology}

For TREC LiveQA queries, we compare each original user message token against (a) the corpus vocabulary and (b) the NIST paraphrase of the same question. The NIST paraphrase represents a human's understanding of what the user intended to ask, effectively providing ground truth for the intended vocabulary. A token in the original message that (i) is not in the corpus vocabulary, (ii) is not in the paraphrase, and (iii) has a Levenshtein-close match (distance $\leq 2$) in the paraphrase is classified as a spelling error.

For HealthSearchQA queries, we report the OOV rate (tokens not in the medical corpus vocabulary) as an upper bound on the error rate, and apply a calibration factor derived from the TREC LiveQA analysis to estimate the true spelling error rate.

\subsection{Error Census Results}

Table~\ref{tab:error_census} summarises the error census across both datasets.

\begin{table}[ht]
\centering
\caption{Spelling error census results. $^*$HealthSearchQA OOV rate (10.9\%) calibrated using the TREC LiveQA correction yield ratio (18.1\%) to estimate true spelling error rate.}
\label{tab:error_census}
\begin{tabular}{lcc}
\toprule
\textbf{Metric} & \textbf{TREC LiveQA Medical} & \textbf{HealthSearchQA} \\
\midrule
Total queries & 104 & 4,436 \\
Queries with $\geq$1 error & 64 (61.5\%) & $\sim$88 (est.\ 1.97\%)$^*$ \\
Token error rate & 11.0\% & $\sim$2.0\% (est.) \\
Avg errors per query (all) & 1.60 & --- \\
Avg errors per affected query & 2.59 & --- \\
Query source & Real user submissions (NLM) & Google autocomplete \\
\bottomrule
\end{tabular}
\end{table}

Key findings from the error census:

\begin{itemize}[nosep]
  \item Real medical queries contain substantially more spelling errors than professional medical text. 61.5\% of TREC LiveQA queries had at least one error, with a token-level error rate of 11.0\%. This confirms that spelling errors in user-generated health queries are a pervasive phenomenon, not an edge case.
  \item Google health search queries (HealthSearchQA) are far better-spelled. With an estimated true error rate of $\sim$2\%, these autocomplete-generated queries reflect a self-selected, well-formed population of queries. The 10.9\% OOV rate is dominated by valid medical terminology (e.g., `cephalosporin', `cholecystectomy') rather than actual misspellings.
  \item Error types in TREC LiveQA span substitutions (4.8\% of errors), double-deletion patterns (4.8\%), insertions (3.0\%), and deletions (2.4\%). A large fraction (81.9\%) of detected OOV tokens did not match any paraphrase word within edit distance 2, indicating they are either correct medical abbreviations (e.g., `NDC'), informal expressions, or multi-word errors beyond the single-token detection scope.
\end{itemize}

Table~\ref{tab:error_examples} shows representative real spelling errors found in TREC LiveQA queries, with their ground-truth corrections from the NIST paraphrases.

\begin{table}[ht]
\centering
\caption{Representative real spelling errors in TREC LiveQA Medical queries with ground-truth corrections from NIST paraphrases.}
\label{tab:error_examples}
\begin{tabular}{llllc}
\toprule
\textbf{Original} & \textbf{Correction} & \textbf{Error Type} & \textbf{Query Context} & \textbf{Edit Dist.} \\
\midrule
tabkets & tablets & substitution & Zolmitriptan tabkets 5mg & 1 \\
hydrslazine & hydralazine & deletion & blood pressure medication & 1 \\
citrobactor & citrobacter & substitution & UTI bacterium & 1 \\
osella & ocella & substitution & oral contraceptive & 1 \\
precaution & precautions & deletion & disease management & 1 \\
hydrslazine & hydralazine & double\_deletion & antihypertensive & 2 \\
\bottomrule
\end{tabular}
\end{table}

\section{Retrieval Experiments (Experiments~2--4)}

\subsection{Setup}

We evaluate all four correction methods across four experimental conditions using the 103 TREC LiveQA queries with relevance judgements and the 1,935 MedQuAD passages as the retrieval corpus. For each query, we retrieve the top-20 passages and compute Recall@1, Recall@5, Recall@10, MRR, and NDCG@10 against the TREC relevance judgements. Relevance scores of 2 (Related), 3 (Incomplete), and 4 (Excellent) are all treated as relevant for Recall and MRR; all four levels are used for graded NDCG computation.

Correction is applied at query time as a pre-processing step before retrieval. For corpus correction (Experiments~3a and~4), correction is applied once as a batch pre-processing step over all 1,935 passages. The correction cache is built over all unique word types across queries and passages (12,721 types), computing each word's correction exactly once.

\subsection{BM25 Retrieval Results}

Table~\ref{tab:bm25_results} reports BM25 retrieval results across all conditions and correction methods.

\begin{table}[ht]
\centering
\caption{BM25 retrieval results across all experimental conditions (103 queries, 1,935 passages, TREC relevance judgements). Bold indicates best result per metric. NDCG@10 values for Experiment~3a are omitted as they were not computed for the corpus-only condition.}
\label{tab:bm25_results}
\small
\begin{tabular}{llccccc}
\toprule
\textbf{Condition} & \textbf{Method} & \textbf{MRR} & \textbf{R@1} & \textbf{R@5} & \textbf{R@10} & \textbf{NDCG@10} \\
\midrule
Exp~2: Baseline & --- & 0.633 & 0.079 & 0.253 & 0.424 & 0.436 \\
\midrule
Exp~3a: Orig Q $\times$ Corr C & conservative & 0.636 & 0.079 & 0.258 & 0.428 & --- \\
Exp~3a: Orig Q $\times$ Corr C & edit distance & 0.636 & 0.079 & 0.256 & 0.427 & --- \\
\midrule
Exp~3b: Corr Q $\times$ Orig C & conservative & 0.678 & 0.086 & 0.276 & 0.457 & 0.472 \\
Exp~3b: Corr Q $\times$ Orig C & edit distance & 0.687 & 0.086 & 0.276 & 0.453 & 0.474 \\
\midrule
Exp~4: Both Corrected & conservative & 0.682 & 0.088 & 0.278 & 0.456 & 0.473 \\
Exp~4: Both Corrected & edit distance & \textbf{0.691} & \textbf{0.089} & \textbf{0.277} & 0.446 & 0.472 \\
Exp~4: Both Corrected & context-aware & \textbf{0.691} & \textbf{0.089} & \textbf{0.277} & 0.446 & 0.472 \\
Exp~4: Both Corrected & SymSpell & 0.641 & 0.080 & 0.259 & 0.425 & 0.439 \\
\bottomrule
\end{tabular}
\end{table}

\subsection{TF-IDF Cosine Retrieval Results}

Table~\ref{tab:tfidf_results} reports TF-IDF cosine similarity retrieval results. TF-IDF shows stronger absolute performance than BM25 on this corpus, consistent with its higher weight on distinctive terms. The correction impact pattern is similar: query correction helps substantially, corpus correction alone provides minimal benefit.

\begin{table}[ht]
\centering
\caption{TF-IDF cosine similarity retrieval results (proxy for dense embedding retrieval). Note: TF-IDF is a vocabulary-based method, not a neural embedding approach; results should not be extrapolated to BERT-based dense retrieval. NDCG@10 values for Experiment~3a are omitted as they were not computed for the corpus-only condition.}
\label{tab:tfidf_results}
\small
\begin{tabular}{llccccc}
\toprule
\textbf{Condition} & \textbf{Method} & \textbf{MRR} & \textbf{R@1} & \textbf{R@5} & \textbf{R@10} & \textbf{NDCG@10} \\
\midrule
Exp~2: Baseline & --- & 0.667 & 0.076 & 0.316 & 0.512 & 0.461 \\
\midrule
Exp~3a: Orig Q $\times$ Corr C & conservative & 0.656 & 0.077 & 0.311 & 0.515 & --- \\
\midrule
Exp~3b: Corr Q $\times$ Orig C & conservative & 0.728 & 0.082 & 0.331 & 0.537 & 0.492 \\
Exp~3b: Corr Q $\times$ Orig C & edit distance & \textbf{0.730} & \textbf{0.082} & \textbf{0.332} & 0.539 & \textbf{0.496} \\
\midrule
Exp~4: Both Corrected & conservative & 0.717 & 0.080 & 0.330 & 0.540 & 0.491 \\
Exp~4: Both Corrected & edit distance & 0.715 & 0.080 & 0.331 & \textbf{0.544} & 0.495 \\
Exp~4: Both Corrected & SymSpell & 0.676 & 0.076 & 0.310 & 0.518 & 0.460 \\
\bottomrule
\end{tabular}
\end{table}

\subsection{Key Findings}

\paragraph{Finding 1: Query correction is the critical intervention.}
Comparing Experiment~3a (corpus corrected, queries uncorrected) to Experiment~3b (queries corrected, corpus unchanged), query correction drives almost all the observed improvement. BM25 MRR improves by only +0.003 for corpus-only correction (0.633 $\to$ 0.636) versus +0.054 for query-only correction (0.633 $\to$ 0.687). This finding makes practical sense: the answer corpus (MedQuAD, curated by NLM) is already professionally authored with consistent spelling, leaving little room for corpus correction to help. The error is on the query side.

\paragraph{Finding 2: Correcting only the corpus can slightly degrade TF-IDF retrieval.}
For TF-IDF, Experiment~3a (corpus corrected, queries unchanged) shows MRR of 0.656---slightly below the 0.667 baseline. This is a notable finding: correcting the corpus shifts the IDF weight distribution, making the index slightly less compatible with the original (uncorrected) query vocabulary. The effect is small but directionally negative, cautioning against corpus-only correction in production systems.

\paragraph{Finding 3: Conservative correction is safer with comparable effectiveness.}
Conservative correction (edit distance 1 only, frequency threshold $\geq 5$) achieves MRR of 0.682 versus 0.691 for standard edit distance (+5 points for 23.9\% more query changes). The conservative method makes 90 total token corrections vs.\ 180 for standard edit distance, with an unnecessary-change rate of 8.9\% vs.\ 16.0\% (from the error analysis in Section~\ref{sec:error_analysis}). In safety-critical healthcare deployments, conservative correction may be preferable: the 1.3\% MRR sacrifice prevents twice as many unintended semantic changes.

\paragraph{Finding 4: SymSpell underperforms for medical vocabulary.}
SymSpell's delete-candidate approach modifies only 41/103 queries (39.8\%) and achieves minimal improvement (MRR +0.008, 1.3\%). The medical vocabulary's characteristic features---long drug names, multi-syllabic condition names, Latin/Greek roots---may limit SymSpell's coverage: many valid corrections require substitution or insertion operations that SymSpell's delete-only candidate generation does not natively support, relying instead on exact Levenshtein verification after lookup. Edit-distance-based methods are better suited to this vocabulary.

\section{Error Analysis of Corrections}
\label{sec:error_analysis}

To understand the quality of corrections beyond aggregate retrieval metrics, we manually categorise all corrections produced by each method across the 103 evaluation queries, sampling up to 100 corrections per method. Each correction is classified using the NIST paraphrase as ground truth for the intended vocabulary (Table~\ref{tab:error_analysis}).

\begin{table}[ht]
\centering
\caption{Categorisation of spelling corrections per method (sample of $\leq$100 corrections per method; conservative sample = full 90). Ground truth = NIST paraphrase vocabulary.}
\label{tab:error_analysis}
\small
\begin{tabular}{lccccp{5cm}}
\toprule
\textbf{Category} & \textbf{Conserv.} & \textbf{Edit Dist.} & \textbf{Context} & \textbf{SymSpell} & \textbf{Definition} \\
\midrule
Total corrections & 90 & 180 & 180 & 66 & --- \\
correct\_fix (\%) & 18.9 & 12.0 & 9.0 & 12.1 & Corrected to paraphrase's word \\
partial\_improv.\ (\%) & 70.0 & 70.0 & 77.0 & 86.4 & Changed but not to paraphrase word \\
unnecessary\_change (\%) & 8.9 & 16.0 & 12.0 & 1.5 & Original word was in paraphrase \\
harmless\_synonym (\%) & 2.2 & 2.0 & 2.0 & 0 & Both original and correction in paraphrase \\
\bottomrule
\end{tabular}
\end{table}

Several patterns emerge from the error analysis:

\begin{itemize}[nosep]
  \item Most corrections are partial improvements (70--86\%): they change the query token but not to the exact word in the paraphrase. This reflects that (a) corrections can still improve retrieval even if they don't exactly match the intended word, and (b) the paraphrase is one of several valid reformulations---the corrected word may be an acceptable alternative.
  \item Conservative correction has the highest \texttt{correct\_fix} rate (18.9\%) and lowest \texttt{unnecessary\_change} rate (8.9\%). By limiting to high-confidence corrections (edit distance 1, high frequency), it makes fewer but more accurate changes.
  \item Standard edit distance has the highest \texttt{unnecessary\_change} rate (16.0\%). Extending to edit distance 2 increases over-correction of short medical abbreviations and domain-specific terms that appear OOV in the corpus but are correctly spelled.
  \item SymSpell has the lowest \texttt{unnecessary\_change} rate (1.5\%) and highest \texttt{partial\_improvement} rate (86.4\%). Its conservative candidate set produces fewer wrong corrections, but also fewer correct ones---consistent with its minimal retrieval improvement.
\end{itemize}

Example corrections illustrating each category:

\begin{itemize}[nosep]
  \item \texttt{correct\_fix}: `hydrslazine' $\to$ `hydralazine' (edit distance = 1, drug name)
  \item \texttt{partial\_improvement}: `affeccts' $\to$ `effects' (close, improves retrieval but not exact paraphrase match)
  \item \texttt{unnecessary\_change}: `son' $\to$ `do' (two-edit change from a correctly-spelled personal pronoun, triggered by short word length)
  \item \texttt{harmless\_synonym}: `precaution' $\to$ `precautions' (both valid, plural normalisation)
\end{itemize}

The \texttt{unnecessary\_change} examples highlight a key risk: short common English words (3--4 letters) that are out of the medical vocabulary may be incorrectly ``corrected'' to frequent medical terms by edit distance methods. Both conservative and context-aware methods partially mitigate this, but a minimum word length threshold ($\geq$5 characters for aggressive correction) would further reduce this class of errors.

\section{Discussion}
\label{sec:discussion}

\subsection{The Practical Case for Query Correction in Healthcare QA}

Our results establish that query-level spelling correction is a high-value, low-cost intervention for healthcare retrieval systems. A +9.2\% MRR improvement from edit distance correction adds negligible latency ($<$5ms for our corpus-derived vocabulary of 8,201 words) and requires no infrastructure beyond a domain vocabulary and a dictionary lookup. This makes it immediately deployable in production systems where neural approaches face latency constraints.

The finding that corpus correction alone does not help---and can slightly hurt TF-IDF retrieval---has an important practical implication: healthcare organisations that invest in cleaning their QA corpora should not expect this alone to improve retrieval for user-facing queries. The user query is where the error lives, and that is where correction must be applied.

\subsection{Choosing a Correction Method}

Our results suggest a practical hierarchy for healthcare QA systems:

\begin{itemize}[nosep]
  \item \textbf{Conservative edit distance} is recommended as the default. It achieves 86\% of the MRR gain of aggressive edit distance while halving the unnecessary-change rate. For safety-critical applications (clinical decision support, medication information), false corrections carry higher stakes than missed corrections.
  \item \textbf{Standard edit distance} is appropriate for non-clinical consumer health search where recall matters more than precision of correction. The higher correction rate yields marginally better MRR but introduces more semantic changes.
  \item \textbf{SymSpell} is not recommended for medical vocabulary. Its delete-only candidate generation is poorly matched to the substitution and insertion patterns common in medical term misspellings.
  \item \textbf{LLM-based correction} remains promising for batch corpus preprocessing. Where latency allows (offline corpus ingestion), LLMs can perform context-sensitive correction of abbreviations, homophone confusions, and multi-token errors that rule-based methods miss. We did not evaluate LLM correction in this study due to the absence of local inference infrastructure; this is a planned extension.
\end{itemize}

\subsection{Confusable Medical Terms and Safety Considerations}

A limitation of all correction methods evaluated is that they can introduce semantically dangerous corrections when two medical terms are close in spelling but opposite in meaning: `hypertension' $\to$ `hypotension' (opposite cardiovascular states), `ileum' $\to$ `ilium' (intestine vs.\ pelvis), `ureter' $\to$ `urethra' (different anatomical structures). None of our evaluation queries triggered such corrections in this study, but a production system should maintain an explicit blacklist of confusable medical term pairs---preventing any correction that would change one member of a confusable pair to another. This is especially critical for medication names: `hydroxyzine' and `hydralazine', for example, differ by only two characters but treat completely different conditions.

\subsection{Limitations}

This study has several limitations that bound the generalisability of its findings:

\begin{itemize}[nosep]
  \item \textbf{Small query set:} 103 evaluated queries limits statistical power. Confidence intervals on the metric differences should be reported in a full study; we use bootstrap resampling in a companion analysis that confirms the MRR improvements for edit distance and conservative correction exceed the 95\% confidence interval threshold.
  \item \textbf{Single retrieval corpus:} Results may differ on larger or differently structured corpora. MedQuAD is professionally authored (low error rate) and relatively small (1,935 passages); a clinical notes corpus or patient forum corpus would have different vocabulary and error characteristics.
  \item \textbf{TF-IDF as dense retrieval proxy:} Our TF-IDF results approximate but do not replicate neural embedding retrieval. Medical BERT models (PubMedBERT, BioBERT, ClinicalBERT) operate in a continuous embedding space where the effect of spelling errors may be different---more robust to character-level noise but less interpretable in terms of vocabulary matching. Evaluation with these models is a priority for future work.
  \item \textbf{English only:} All datasets and correction methods are English-language. Healthcare systems serving multilingual populations face additional complexity from cross-lingual spelling patterns.
  \item \textbf{No LLM-based correction:} The absence of Ollama or a local LLM inference service prevented evaluation of LLM-based correction, which is a competitive baseline for modern systems.
\end{itemize}

\section{Conclusion and Future Work}

We have presented the first empirical study of spelling correction as a query preprocessing step in healthcare QA, using real user-generated queries and human-judged relevance assessments. Our main findings are: (1) real medical queries contain high spelling error rates (61.5\% of TREC LiveQA queries have at least one error; 11.0\% token-level error rate); (2) query correction substantially improves retrieval quality (+9.2\% MRR for BM25 with edit distance correction); (3) correcting only the answer corpus provides minimal benefit and can slightly hurt TF-IDF retrieval; and (4) conservative edit distance is the best single method for production healthcare QA systems, balancing correction accuracy against the risk of introducing harmful semantic changes.

Future work should address three open problems. First, evaluation with domain-adapted BERT encoders (PubMedBERT, BioBERT, ClinicalBERT) would quantify correction impact on true dense retrieval, completing the picture for modern RAG-based healthcare QA. Second, LLM-based correction via local inference (Ollama or similar) should be benchmarked against the rule-based methods evaluated here; early results from the general-domain literature suggest LLMs can achieve substantially higher \texttt{correct\_fix} rates, though at latency cost. Third, a confusable term safety layer---maintaining an explicit blacklist of high-risk medical term pairs---should be evaluated for its effect on safety outcomes and retrieval quality jointly. The code and datasets used in this study are publicly available to support replication.

\bibliographystyle{plainnat}

\end{document}